\documentclass{article}
\usepackage{arxiv}
\usepackage[utf8]{inputenc} 
\usepackage[T1]{fontenc}    
\usepackage[hidelinks]{hyperref}       
\usepackage{url}            
\usepackage{booktabs}       
\usepackage{amsfonts}       
\usepackage{nicefrac}       
\usepackage{microtype}      
\usepackage{graphicx}
\graphicspath{ {./images/} }

\makeatletter
\renewcommand{\@seccntformat}[1]{\csname the#1\endcsname.\enspace}
\renewcommand{\today}{}
\makeatother

\date{}  

\title{KnowThyself: An Agentic Assistant for LLM Interpretability}

\author{
Suraj Prasai\textsuperscript{1},
Mengnan Du\textsuperscript{2},
Ying Zhang\textsuperscript{1},
Fan Yang\textsuperscript{1} \\
\textsuperscript{1}Wake Forest University\\
\textsuperscript{2}New Jersey Institute of Technology \\
\texttt{\{prass25, zhangyi, yangfan\}@wfu.edu}, \texttt{mengnan.du@njit.edu}
}

\begin{document}
\maketitle

\begingroup
\renewcommand\thefootnote{}%
\footnote{\textit{This paper has been accepted for publication at the Demonstration Track of the 40th AAAI Conference on Artificial Intelligence (AAAI'26). 
This is the preprint version. The final published version will appear in the AAAI-26 proceedings.}}%
\addtocounter{footnote}{-1}%
\endgroup

\begin{abstract}
We develop \texttt{KnowThyself}, an agentic assistant that advances large language model (LLM) interpretability. Existing tools provide useful insights but remain fragmented and code-intensive. \texttt{KnowThyself} consolidates these capabilities into a chat-based interface, where users can upload models, pose natural language questions, and obtain interactive visualizations with guided explanations. At its core, an orchestrator LLM first reformulates user queries, an agent router further directs them to specialized modules, and the outputs are finally contextualized into coherent explanations. This design lowers technical barriers and provides an extensible platform for LLM inspection. By embedding the whole process into a conversational workflow, \texttt{KnowThyself} offers a robust foundation for accessible LLM interpretability.
\end{abstract}

\section{Introduction}

Large language models (LLMs) have attracted significant attention for their impressive capabilities in language understanding, reasoning, and problem solving~\cite{naveed2025comprehensive}. However, their black-box nature makes it difficult to interpret internal decision processes, raising concerns about transparency, trust, and accountability~\cite{zhao2024explainability,huang2024trustllm}. Although recent research has sought to explain LLM behavior, progress in interpretability has largely lagged behind the rapid pace of LLM development. 

Existing LLM interpretability approaches include attribution methods that assign importance scores to tokens, samples, or hidden states~\cite{li2023survey,lee2025llm}, as well as mechanistic analyses of attention heads, neurons, or circuits~\cite{dunefsky2024transcoders,gantla2025exploring}. While these approaches provide valuable insights, they remain isolated, difficult to use, and require substantial technical expertise. Such shortcomings create a gap between cutting-edge interpretability research and its practical accessibility in real-world settings~\cite{wu2024usable,singh2024rethinking}. For LLM practitioners, significant barriers to accessing interpretability persist, since current platforms neither support conversational exploration nor provide interactive, well-grounded explanations. These barriers slow the democratization of interpretability and limit the pace at which broader audiences can engage with emerging interpretation techniques. 

To bridge this gap, we introduce \texttt{KnowThyself}, an agentic platform that unifies interpretability tools within an accessible and extensible framework. Our system integrates multi-agent \textbf{orchestration}, modular \textbf{architecture}, and interactive \textbf{visualization} into a single \textit{conversational} workflow. Unlike existing fragmented tools, \texttt{KnowThyself} allows users to upload models, pose natural language questions, and obtain both visual outputs and explanatory responses without writing code. Our main contributions include: (i) a multi-agent orchestration framework that coordinates a broad range of interpretation tasks, enabling flexible routing and producing coherent explanations; (ii) a modular architecture that encapsulates different methods as independent agents, supporting seamless integration of new tools and scalable extension in future; and (iii) an interactive visualization interface that presents outputs with natural language explanations, significantly lowering barriers to effective model inspection.  

\begin{figure*}[t]
\centering
\includegraphics[width=0.99\textwidth]{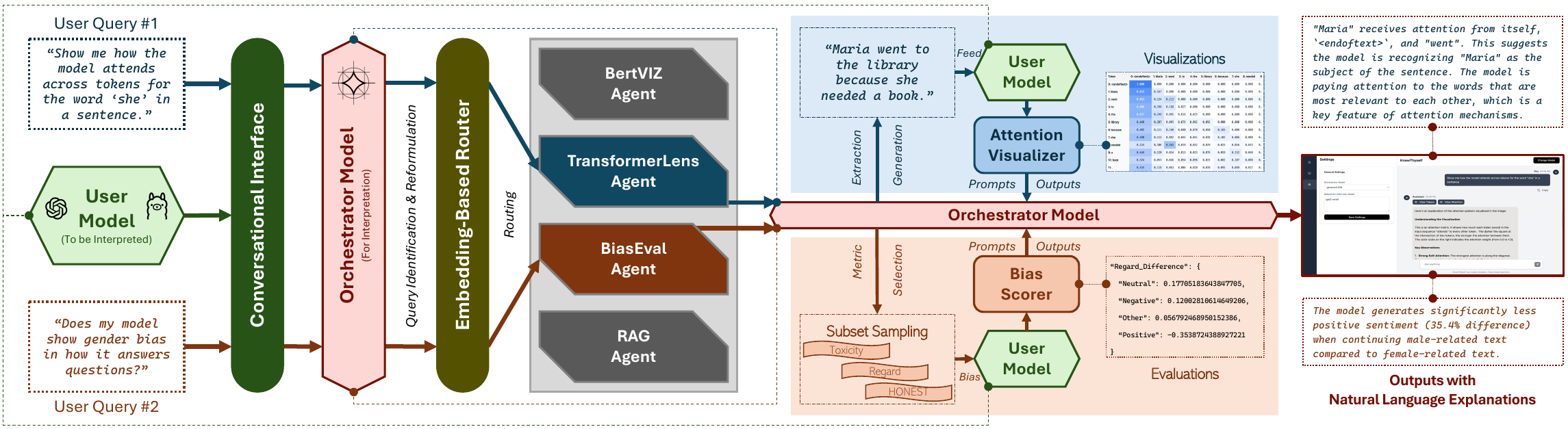}
\caption{The agentic pipeline of \texttt{KnowThyself} for two demonstrative case studies on \textit{token attribution} and \textit{bias evaluation}.}
\label{fig:interface_and_model_view}
\end{figure*}

\section{System Overview}

\texttt{KnowThyself} is an agentic platform that unifies the interpretation process into a conversational workflow. Rather than requiring users to operate standalone libraries, it introduces an abstraction layer that translates natural language queries into tool invocations and returns guided explanations. Our system consists of four components: an \textbf{Orchestrator LLM} for reformulation, an \textbf{Agent Router} for selection, \textbf{Specialized Agents} for analysis, and a \textbf{Conversational Interface} for interaction. The illustrative pipeline of \texttt{KnowThyself} is shown in Figure~\ref{fig:interface_and_model_view}. 

\paragraph{Orchestrator LLM.}
The orchestrator serves as a supervisory model that manages user interactions and directs the interpretation process. It reformulates queries, generates necessary subtasks (e.g., \textit{sentence synthesis} or \textit{tool selection}), and contextualizes intermediate results. Finally, it produces coherent natural language explanations, ensuring that complex visualizations or bias metrics remain understandable.

\paragraph{Agent Router.}
The router dispatches queries to specialized agents using embedding-based similarity search to match user intent with agent descriptions. This ensures alignment between queries and tool capabilities while maintaining efficiency. As the system scales, it can be augmented with LLM-based routing for adaptability in complex cases. 

\paragraph{Specialized Agents.}
Each agent encapsulates an interpretation method as a modular plug-in. The current system integrates four agents: (i) \texttt{BertViz}~\cite{bertviz} for attention visualization, (ii) \texttt{TransformerLens}~\cite{transformerlens} for analyzing fine-grained layer- and head-level activations, (iii) \texttt{RAG} explainer that grounds responses in domain literature, and (iv) \texttt{BiasEval} which assesses safety and demographic disparities using \textit{toxicity}~\cite{toxicityPrompts}, \textit{regard}~\cite{regardPaper}, and \textit{HONEST}~\cite{honestScore} scores.

\paragraph{Conversational Interface.}
The chat interface allows users to upload models, pose questions in natural language, and examine results with interactive visualizations, making exploration accessible without requiring technical expertise.

\section{Implementation}

We implement the system with LangGraph~\cite{langgraph}, modeling as a directed graph of agents over a shared state. Query routing relies on embedding-based similarity search with the Ollama-hosted \texttt{nomic-embed-text} model~\cite{embeddingModel}, while orchestration is managed by Gemma3-27B~\cite{gemma3}. For user models, we pre-include GPT-2~\cite{gpt2}, BERT~\cite{DBLP}, and LLaMA2-13B~\cite{llama2} for demonstration. Large models are served through Ollama for efficient hosting, and the system is able to run locally when resources permit, ensuring secure analysis without third-party APIs. 

Different interpretation tools require distinct dependencies, encapsulated within respective agents. For instance, \texttt{TransformerLens} relies on \textit{HookedTransformer}, while \texttt{BertViz} builds on \textit{HuggingFace Transformers}~\cite{transformers}. For bias analysis, \texttt{BiasEval} prompts models with Real Toxicity Prompts~\cite{toxicityPrompts}, BOLD~\cite{boldRegardDataset}, and HONEST~\cite{honestScore} datasets, reporting \textit{toxicity}, \textit{regard}, and \textit{HONEST} scores. The \texttt{RAG} agent indexes documents and applies FAISS~\cite{faiss} for similarity search, retrieving information that the Orchestrator LLM incorporates as context for grounded explanations. By isolating these dependencies, new tools can be integrated without disrupting the system. Such modular design supports independent development while ensuring the platform remains extensible. 

\section{Use Cases}
\texttt{KnowThyself} supports practical scenarios where interpretability of LLMs is a central concern. 
As shown in Figure~\ref{fig:interface_and_model_view}, a user may upload a LLaMA2 checkpoint and ask, ``\textit{Show me how the model attends across tokens for the word ‘she’ in a sentence.}''. The Agent Router selects \texttt{TransformerLens}, and the Orchestrator supplies required inputs by synthesizing a sentence (e.g., ``\textit{Maria went to the library because she needed a book.}'') when no input is provided. \texttt{TransformerLens} then computes attention maps and returns an interactive visualization, which the Orchestrator contextualizes into a coherent explanation. 
In the same session, the user may ask, ``\textit{Does my model show gender bias in how it answers questions?}''. The Orchestrator identifies this as a new task rather than a follow-up, and the Agent Router further selects \texttt{BiasEval} that queries the Orchestrator to choose the relevant submodule (e.g., \textit{regard}), samples prompts from the BOLD dataset, runs them on the user model, and computes the scores. Finally, the Orchestrator summarizes the results and presents them to the user. Overall, \texttt{KnowThyself} conducts the interpretation process within a conversational flow, allowing users to move seamlessly between tasks while receiving clear explanations and interactive visualizations in context.

\section{Conclusion and Future Work}
We present \texttt{KnowThyself}, a conversational multi-agent platform for LLM interpretability. Our system streamlines the interpretability through a conversational workflow, integrates interactive visualizations with literature-grounded explanations, and adopts a modular architecture that enables new methods to be incorporated without altering core components. By lowering technical barriers, \texttt{KnowThyself} empowers LLM practitioners to engage with model interpretability issues more effectively without certain expertise. Nonetheless, the current implementation integrates only a limited set of tools, requires additional engineering to adapt non-modular libraries, and supports text inputs exclusively. Future work will broaden tool coverage, extend support to multimodal models, improve routing precision for overlapping tasks, and introduce richer visualization capabilities for deeper and more transparent interpretive insights.

\section*{Reproducibility and Acknowledgments}
To promote reproducibility and future extensions, the implementation of \texttt{KnowThyself} is publicly available at: \texttt{https://github.com/spygaurad/KnowThyself}. The authors gratefully acknowledge Jiru Xu, Xuansheng Wu, Shushi Hong, Chris (Tian) Xia, and Ninghao Liu for their valuable discussions, technical feedback, and constructive contributions during the development of this work. The authors also thank the anonymous reviewers for their insightful comments that helped improve the clarity and quality of the manuscript. This research was supported in part by the U.S. National Science Foundation under Grant IIS2451480. 

\bibliographystyle{unsrt}  
\bibliography{references}
\end{document}